\documentclass[10pt,twocolumn,letterpaper]{article}

\usepackage[pagenumbers]{cvpr} %

\usepackage{graphicx}
\usepackage{amsmath}
\usepackage{amssymb}
\usepackage{booktabs}
\usepackage{bbold}
\usepackage{multirow}
\usepackage[dvipsnames]{xcolor}

\newcommand{\vct}[1]{\boldsymbol{#1}}

\newcommand{\vw}{\vct{w}}
\newcommand{\itared}[1]{\textbf{\textit{\textcolor{black}{#1}}}}
\newcommand{\boblu}[1]{\textbf{\textcolor{black}{#1}}}

\newcommand{\down}{\textcolor{green}{$\downarrow$}}
\newcommand{\method}[1]{\textsc{#1}}

\newcommand{\DS}{D_{\mathcal{S}}{}}
\newcommand{\DT}{D_{\mathcal{T}}{}}

\usepackage[pagebackref,breaklinks,colorlinks]{hyperref}

\usepackage[capitalize]{cleveref}
\crefname{section}{Sec.}{Secs.}
\Crefname{section}{Section}{Sections}
\Crefname{table}{Table}{Tables}
\crefname{table}{Tab.}{Tabs.}

\begin{document}

\title{Burn After Reading: \\Online Adaptation for Cross-domain Streaming Data}

\author{
$\textbf{Luyu Yang}^{1}, \textbf{Mingfei Gao}^{2}, \textbf{Zeyuan Chen}^{2}, \textbf{Ran Xu}^{2},$ \\ $\textbf{Abhinav Shrivastava}^{1}, \textbf{Chetan Ramaiah}^{2}$\\
\vspace{-0.15in}\\
$^{1}\text{University of Maryland}$\quad \quad
$^{2}\text{Salesforce Research}$
}
\maketitle

\begin{abstract}
In the context of online privacy, many methods propose complex privacy and security preserving measures to protect sensitive data. In this paper we argue that: not storing any sensitive data is the best form of security. Thus we propose an online framework that ``burns after reading'', i.e. each online sample is immediately deleted after it is processed. Meanwhile, we tackle the inevitable distribution shift between the labeled public data and unlabeled private data as a problem of unsupervised domain adaptation. Specifically, we propose a novel algorithm that aims at the most fundamental challenge of the online adaptation setting--the lack of diverse source-target data pairs. Therefore, we design a \textbf{Cro}ss-\textbf{Do}main \textbf{Bo}otstrapping approach, called \method{\textbf{CroDoBo}}, to increase the combined diversity across domains. Further, to fully exploit the valuable discrepancies among the diverse combinations, we employ the training strategy of multiple learners with co-supervision. \method{CroDoBo} achieves state-of-the-art online performance on four domain adaptation benchmarks.
\end{abstract}

\section{Introduction}
With the onslaught of the pandemic, the internet has become an even more ubiquitous presence in all of our lives. Living in an enormous web connecting us to each other, we now face a new reality: it is very hard to escape one's past on the Internet now that every photo, status update, and tweet lives forever in the cloud~\cite{rosen2011right,tirosh2017reconsidering,villaronga2018humans,garg2020formalizing}.
Moreover, recommender systems that actively explore the user data~\cite{guo2021embedding,zhang2021alicg} for data-driven algorithms have encouraged the debate about the right to privacy over convenience.
Fortunately, we have \emph{the Right to Be Forgotten} (RTBF), which gives individuals the right to ask organizations to delete their personal data.
Recently, many solutions~\cite{ryffel2018generic,wang2019privacy,zhang2019deeppar,ziller2021medical} have been proposed that try to preserve privacy in the context of deep learning, mostly focused on the Federated Learning~\cite{wei2020federated,kaissis2020secure,xu2019hybridalpha,truex2019hybrid}. Federated Learning allows asynchronous update of multiple nodes, in which sensitive data is stored only on a few specific nodes. However, recent studies~\cite{zhu2020deep,jin2021cafe,xu2020information} show that private training data can be leaked through the gradients sharing mechanism deployed in distributed models. In this paper, we argue that: \emph{not storing any sensitive data is the best form of security}.

The best form of security requires us to delete the user data after use, which necessitates an online framework. However, existing online learning frameworks~\cite{sahoo2017online,perozzi2014deepwalk,liu2017survey} cannot meet this need without addressing the distribution shift from public data, \textit{i.e.} source domain, to the private user data, \textit{i.e.} target domain. Therefore, in this paper we propose an online domain adaptation framework in which the target domain streaming data is deleted immediately after adapted. The task that is seemingly an extended setting of unsupervised domain adaptation (UDA), however, cannot simply be solved by the online implementation of the offline UDA methods. We explain the reason with a comprehensive analysis of the existing domain adaptation methods. To begin with, existing offline UDA methods rely heavily on the rich combinations of cross-domain mini-batches that gradually adjust the model for adaptation~\cite{ganin2016domain,saito2018maximum,lee2019drop,tzeng2017adversarial,yang2020curriculum,peng2019moment,zhang2019bridging,vu2019advent,shu2018dirt}, which the online streaming setting cannot afford to provide. In particular, many domain adversarial-based methods~\cite{ganin2016domain,wang2018unsupervised,lafarge2017domain,hu2020discriminative,chen2020domain,pei2018multi} depend on a slowly annealing adversarial mechanism that requires discriminating large number of source-target pairs to achieve the adaptation. Recently, state-of-the-art offline methods~\cite{kang2019contrastive,liang2020we,liang2021domain} show promising results by exploiting target-oriented clustering, which requires an offline access to the entire target domain. Therefore, the online UDA task needs new solutions to succeed at scarcity of the data from target domain.

\begin{figure*}[t]
    \centering
    \includegraphics[width=0.98\linewidth]{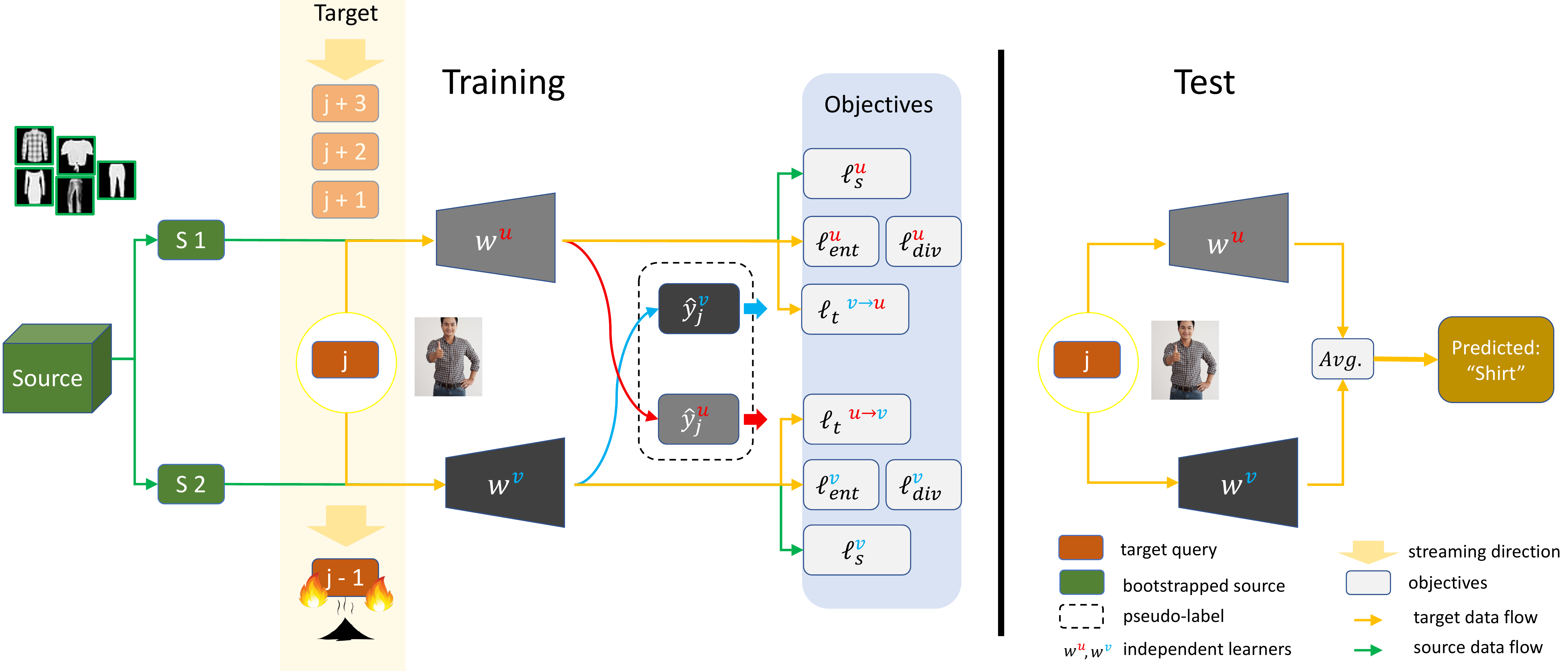}
    \caption{\small The pipeline of the proposed \method{CroDoBo} framework. \method{CroDoBo} adapts from a public source domain to a streaming online target domain. At each timestep, only the current \emph{$j$-th query} is available from the target domain. \method{CroDoBo} bootstraps the source domain batches that combine with the current $j$-th query in order to increase the cross-domain data diversity. The learners $w^u$ and $w^v$ exchange the generated pseudo-labels $\hat{y}_j^{u}$ and $\hat{y}_j^{v}$ as co-supervision. Once the current query is adapted in the training phase, it is tested immediately to make a prediction. Each target query is deleted after tested. Example images of source domain are from Fashion-MNIST~\cite{xiao2017fashion}, adapting to target domain Deep-Fashion~\cite{liu2016deepfashion}. ~\textit{Best viewed in color.}}
    \label{fig:approach_pipeline}
    \vspace{-0.1in}
\end{figure*}

We aim straight at the most fundamental challenge of the online task—the lack of diverse cross-domain data pairs—and propose a novel algorithm based on cross-domain bootstrapping for online domain adaptation. At each online query, we increase the data diversity across domains by bootstrapping the source domain to form diverse combinations with the current target query. To fully exploit the valuable discrepancies among the diverse combinations, we train a set of independent learners to preserve the differences. Inspired by~\cite{yang2020deep}, we later integrate the knowledge of learners by exchanging their predicted pseudo-labels on the current target query to co-supervise the learning on the target domain, but without sharing the weights to maintain the learners' divergence. We obtain more accurate prediction on the current target query by an average ensemble of the diverse expertise of all the learners. We call it \method{\textbf{CroDoBo}}: \textbf{Cr}oss-\textbf{Do}main \textbf{Bo}otstrapping for online domain adaptation, an overview of \method{CroDoBo} is shown in Figure~\ref{fig:approach_pipeline}.

We conduct extensive evaluations on our method, including the classic UDA benchmark \textit{VisDA-C}~\cite{peng2018visda}, a practical medical imaging benchmark \textit{COVID-DA}~\cite{zhang2020covid} and the large-scale distribution shift benchmark \textit{WILDS}~\cite{koh2021wilds} subset \textit{Camelyon}. Moreover, we propose a new adaptation scenario in this paper from \textit{Fashion-MNIST}~\cite{xiao2017fashion} to \textit{DeepFashion}~\cite{liu2016deepfashion}. On all the benchmarks, our method outperforms the state-of-the-art UDA methods that are eligible for the online setting. Further, without the reuse of any target sample, our method achieves comparable performance to the offline setting. We summarize the contributions as follows. (1) We propose an online domain adaptation framework to implement \textit{the right to be forgotten}. (2) We propose a novel online domain adaptation algorithm that achieves new state-of-the-art online results, and comparable results to the offline setting. (3) Despite being simple, the comparable performance to the offline setting suggests that our method is an excellent choice even just for time efficiency.

\section{Related Work}
\noindent \textbf{The Right to Be Forgotten}~\cite{villaronga2018humans,garg2020formalizing,pagallo2019human,tirosh2017reconsidering,byrum2018european}, also referred to as \emph{right to vanish}, \emph{right to erasure} and \emph{courtesy vanishing}, is the right given to each individual to ask organizations to delete their personal data. RTBF is part of the General Data Protection Regulation (GDPR). As a legal document, the GDPR outlines the specific circumstances under which the right applies in \textit{Article 17} GDPR~\footnote{\textit{Article 17} GDPR - Right to be forgotten~\url{https://gdpr.eu/article-17-right-to-be-forgotten/}}. The first item is: \textit{The personal data is no longer necessary for the purpose an organization originally collected or processed it.} Yet, the exercise of this right has become a thorny issue in applications. Politou \etal ~\cite{politou2022right} discussed that the technical challenges of aligning modern systems and processes with the GDPR provisions are numerous and in most cases insurmountable. In ~\cite{politou2018backups} they specifically examined the implications of erasure requests on current backup systems and highlight a number of challenges pertained to the widely known backup standards, data retention policies, backup mediums, search services, and ERP systems~\cite{ali2017erp}. In the context of machine learning, Villaronga \etal~\cite{villaronga2018humans} addressed that the core issue of the AI and Right to Be Forgotten problem is the dearth of interdisciplinary scholarship supporting privacy law and regulation. Graves \etal~\cite{graves2020does} proposed three defense mechanisms against a general threat model to enable deep neural networks to forget sensitive data while maintaining model efficacy. In this paper, we focus on how to obtain model efficacy while erasing data online to protect the user's right to be forgotten.

\noindent \textbf{Online Adaptation to Shifting Domains} was first investigated in Signal Processing~\cite{elliott2000frequency} and later studied in Natural Language Processing~\cite{dredze2008online} and Vision tasks~\cite{qi2008two,jain2011online,mancini2018kitting,gaidon2015online,xu2016hierarchical,csurka2017domain,moon2020multi,delussu2021online}. Jain \etal~\cite{jain2011online} assumed the original classifier output a continuous number of which a threshold gives the class, and reclassify points near the original boundary using a Gaussian process regression scheme. The procedure is presented as a Viola-Jones cascade of classifiers. Moon \etal~\cite{moon2020multi} proposed a four-stage method by assuming a transformation matrix between the source subspace and the mean-target subspace embedded in the Grassmann manifold. The method is designed for handcrafted features. In the context of deep neural network, we argue that one transformation matrix might not be sufficient to describe the correlation between source and target deep representations~\cite{nanni2017handcrafted}. Taufique \etal~\cite{taufique2021conda} approached the task by selectively mixing the online target samples with those that were saved in a buffer. Without a further discussion of which samples can be saved in the buffer, we find this method limited in the exercise of the right to be forgotten.

\noindent \textbf{Active Domain Adaptation}~\cite{rai2010domain,saha2011active,ma2021active,matasci2012svm,chen2021active,persello2012active,deng2018active,prabhu2021active} also benefits the online learning of shifting domains. It bears a different setting: the target domain can actively acquire labeled data online. Rai \etal~\cite{rai2010domain} presented an algorithm that harnessed the source domain data to learn a initializer hypothesis, which is later used for active learning on the target domain. Ma \etal~\cite{ma2021active} allowed a small budget of target data for the categories that appeared only in target domain and presented an algorithm that jointly trains two sub-networks of different learning strategies. Chen \etal~\cite{chen2021active} proposed an algorithm that can adaptively deal with interleaving spans of inputs from different domains by a tight trade-off that depends on the duration and dimensionality of the hidden domains.

\noindent \textbf{Ensemble Methods for Online Learning}~\cite{wang2014resampling,de2016boosting,minku2009impact}
such as bagging and boosting have shown advantages handling \textit{concept drift}~\cite{lu2018learning} and class imbalance, which are common challenges in the online learning task.
MinKu \etal~\cite{minku2009impact} addressed the importance of ensemble diversity to improve accuracy in changing environments and proposed the measurement of ensemble diversity. Han \etal~\cite{han2017branchout} proposed a regularization for online tracking with a subset of branches in the neural network that are randomly selected.
Although online learning and online domain adaptation share similar streaming form of data input, we argue that the two tasks face fundamentally different challenges. For online learning, the challenge is to select the most trustworthy supervisions from the streaming data by differentiating the informative vs. misleading data points, also known as the \textit{stability-plasticity dilemma}~\cite{jaber2013online}. For online domain adaptation, the streaming data of target domain naturally comes unlabeled, and the challenge is the scarcity of supervision. Thus the goal is how to maximize the utilization of the supervision from a different but related labeled source domain.
\section{Approach}

\subsection{Offline vs. Online}
\label{sec:task_compare}
Given the labeled source data $\DS=\{(s_i,y_i)\}_{i=1}^{N_S}$ drawn from the source distribution $p_s(x,y)$, and the unlabeled target data $\DT=\{t_i\}_{i=1}^{N_T}$ drawn from the target distribution $p_t(x,y)$, where $N_S$ and $N_T$ represent the number of source and target samples, both offline and online adaptation aim at learning a classifier that make accurate predictions on $\DT$.
The \emph{offline adaptation} assumes access to every data point in $\DS$ or $\DT$, synchronous~\cite{ganin2016domain,saito2018maximum,shu2018dirt,tzeng2017adversarial,yang2020curriculum} or asynchronous~\cite{liang2020we} domain-wise. The inference on $\DT$ happens after the model is trained on both $\DS$ and $\DT$ entirely. For the \emph{online adaptation}, we assume the access to the entire $\DS$, while the data from $\DT$ arrives in a random streaming fashion of mini-batches $\{T_j=\{t_b\}_{b=1}^B\}_{j=1}^{M_T}$, where $B$ is the batch size, $M_T$ is the total number of target batches. Each mini-batch $T$ is first adapted, tested and then erased from $\DT$ without replacement, as shown in Figure~\ref{fig:approach_pipeline}. We refer each online batch as a \emph{query} in the rest of the paper.

The fundamental challenge of the online task is the limited access to the training data at each inference query, compared to the offline task. Without loss of generality, let's assume there are $10^3$ source and target batches respectively. In an offline setting, the model is tested after training on at most $10^6$ combinations of source-target data pairs, while in an online setting, an one-stream model can see at most $10^3$+$500$ combinations at the $500$-th query.
Undoubtedly, the online adaptation faces a significantly smaller data pool and data diversity, and the training process of the online task suffers from two major drawbacks. First, the model is prone to underfitting~\cite{van2010process} on target domain due to the limited exposure, especially at the beginning of training. Second, due to the erasure of ``seen'' batches, the model lacks the diverse combinations of source-target data pairs that enable the deep network to find the optimal cross-domain classifier~\cite{li2018learning}.
The goal of the proposed method is to minimize these drawbacks of the online setting. We first propose to increase the data diversity by cross-domain bootstrapping, which are preserved in multiple independent learners. Then we fully exploit the valuable discrepancies of these learners by exchanging their expertise on the current target query to co-supervise each other.

\subsection{Proposed Method}
\subsubsection{Cross-domain Bootstrapping for Data Diversity}
\label{sec:bootstrap}
The diversity of cross-domain data pairs is crucial for most prior offline methods~\cite{ganin2016domain,peng2019moment,saito2018maximum} to succeed.
Since the target samples cannot be reused in the online setting, we propose to increase the data diversity across domains by bootstrapping the source domain to form diverse combinations with the current target domain query. Specifically, for each target query $T_j$, we randomly select a set of $K$ mini-batches $\{S_j^k = \{(s_b)_{b=1}^{B}\}\}_{k=1}^{K}$ of the same size from the source domain with replacement. Correspondingly, we define a set of $K$ base learners $\{\vw^k\}_{k=1}^{K}$. At each iteration, a learner $\vw^k$ makes prediction for query $T_j$ after trained on $\{T_j, S_j^k\}$, and updates via
\begin{align}
    \vw^k \leftarrow ~&\vw^k - \eta\left(\nabla\mathcal{L}(\vw^k, \{T_j, S_j^k\})\right) \nonumber\\
    p_{j}^{k} =~&p\left(c|T_j; \vw^k\right) \
\end{align}
where $\eta$ is the learning rate, $c$ is the number of classes, $p_{j}^{k}$ is the predicted probability by the $k$-th learner, and $\mathcal{L}(,)$ is the objective function. The predicted class for $T_j$ is the average of $K$ predictions of the base learners. We justify our design choice from the perspective of uncertainty estimation in the following discussion.

\noindent \textbf{Theoretical Insights} As mentioned in \cref{sec:task_compare}, we aim at the best estimation of the current target query. We first consider a single learner situation.
At the $j$-th query, the learner faces a fundamental trade-off:
by minimizing the uncertainty of the $j$-th query, the learner can attain the best current estimation. Yet the risk of fully exploring the uncertainty is to spoil the existing knowledge from the previous $j$-1 target domain queries. However, if we don't treat the uncertainty, the single observation on $j$-th query is less informative for current query estimation. Confronting the dilemma, we should not ignore that the uncertainty captures the variability of a learner's posterior belief which \textit{can} be resolved through statistical analysis of the appropriate data~\cite{burnetas1997optimal,osband2016risk}. This gives us hope for a more accurate model via uncertainty estimation. One popular suggestion for resolving uncertainty is to use \textit{Dropout}~\cite{gal2016dropout, gal2017concrete,saito2017adversarial} sampling, where individual neurons are independently set to zero with a probability. 
As a sampling method on the neurons, \textit{Dropout} works in a similar form of \textit{bagging}~\cite{warde2013empirical,schmitz2014tactile} of multiple decision trees. It might equally reduce the overall noise of the network regardless of domain shift but it does not address the problem of our task, which is the lack of diverse cross-domain combinations.

Alternatively, we employ another pragmatic approach \textit{Bootstrap} for uncertainty estimation on the target domain that offsets the source dominance. With the scarcity of target samples, we propose to bootstrap source-target data pairs for a more balanced cross-domain simulation. At high-level, the bootstrap simulates multiple realizations of a specific target query given the diversity of source samples. Specifically, the bootstrapped source approximate a distribution over the current query $T_j$ via the bootstrap.

The bootstrapping brings multi-view observations on a single target query by two means. First, given $K$ sampling subsets from $\DS$, let $\mathcal{F}$ be the ideal estimate of $T_j$, $\hat{\mathcal{F}}$ be the practical estimate of the dataset, and $\hat{\mathcal{F}}^{*}$ be the estimate from a bootstrapped source paired with the target query,
\begin{align}
    \hat{\mathcal{F}}^{*} = K^{-1}\sum_{k=1}^{K}\hat{\mathcal{F}}^{*}_{k}
\end{align}
will be the average of the multi-view $K$ estimates. Second, besides the learnable parameters, the \textit{Batch-Normalization} layers of $K$ learners generate result in a set of different means and variances $\{\mu_k, \sigma_k\}_{k=1}^{K}$ that serve as $K$ different initializations that affects the learning of $\hat{\mathcal{F}}^{*}$.

\subsubsection{Exploit the Discrepancies via Co-supervision}
\label{sec:exchange}
After the independent learners have preserved the valuable discrepancies of cross-domain pairs, the question now is how to fully exploit the discrepancies to improve the online predictions on the target queries. On one hand, we want to integrate the learners' expertise into one better prediction on the current target query, on the other we hope to maintain their differences. Inspired by~\cite{yang2020deep}, we train the $K$ learners jointly by exchanging their knowledge on the target domain as a form of co-supervision. Specifically, the $K$ learners are trained independently with bootstrapped source supervision, but they exchange the pseudo-labels generated for target queries.
We followed the \textit{FixMatch} to compute pseudo-labels on the target domain. In this paper, we consider $K$=2 for simplicity, we denote the learners as $\vw^{u}$ and $\vw^{v}$ for $k=1$, $k=2$ respectively.

Given the current target query $T_j$, the loss function $\mathcal{L}$ consists a supervised loss term $\ell_s$ from the source domain with the bootstrapped samples, and a self-supervised loss term $\ell_t$ from the target domain with pseudo-labels $\hat{y}_b$ from the peer learner. We denote the cross-entropy between two probability distributions as $\mathcal{H}(;)$.
\begin{align}
\ell_t^{v\rightarrow u} &= B^{-1}\sum_{b=1}^{B} \mathbb{1} \left(p^v_b \geq \tau\right) \mathcal{H}\left(\hat{y}^v_b;p(c|\Tilde{t}_b;\vw^u)\right) \nonumber\\
\ell_t^{u\rightarrow v} &= B^{-1}\sum_{b=1}^{B} \mathbb{1} \left(p^u_b \geq \tau\right) \mathcal{H}\left(\hat{y}^u_b;p(c|\Tilde{t}_b;\vw^v)\right)
\label{eq:fixmatch}
\end{align}
$p^u_b$ and $p^v_b$ are the predicted probabilities of $t_b$ by $\vw^{u}$ and $\vw^{v}$, respectively. $\Tilde{t}_b$ is a strongly-augmented version of $t_b$ using \textit{Randaugment}~\cite{cubuk2020randaugment}, and $\tau$ is the threshold for pseudo-label selection. To further exploit the supervision from the limited target query, from $p^u_b$ and $p^v_b$ we compute a standard entropy minimization term $\ell_\text{ent}$ and a class-balancing diversity term $\ell_\text{div}$ which are widely used in prior domain adaptation works~\cite{vu2019advent,saito2019semi,liang2020we}. Finally, we update the learners by
\begin{align}
    \vw^u \leftarrow \vw^u & - \eta (~\nabla \ell_s (\vw^u, S_j^u) + \nabla\ell_t^{v\rightarrow u} \nonumber\\
    &+ \nabla\ell_\text{ent}(\vw^u, T_j) +  \lambda\nabla\ell_\text{div}(\vw^u, T_j) ) \nonumber\\
     \vw^v \leftarrow \vw^v & - \eta (~ \nabla \ell_s (\vw^v, S_j^v) + \nabla\ell_t^{u\rightarrow v}\\
    &+ \nabla\ell_\text{ent}(\vw^v, T_j) +  \lambda\nabla\ell_\text{div}(\vw^v, T_j)), \nonumber
\end{align}
where $\lambda$ is a hyperparameter that scales the weight of the diversity term.
\section{Experiments}
We consider two metrics for evaluating online domain adaptation methods: \textit{online average accuracy} and \textit{one-pass accuracy}. The online average is an overall estimate of the streaming effectiveness. The one-pass accuracy measures after training on the finite-sample how much the online model has deviated from the beginning~\cite{nakkiran2020deep}. A \textit{one-pass accuracy} much lower than \textit{online average} indicates that the model might have overfitted to the fresh queries, but compromised its generalization ability to the early queries.

\noindent \textbf{Dataset.} We use \textbf{VisDA-C}~\cite{peng2018visda}, a classic benchmark adapting from synthetic images to real. We followed the data split used in prior offline settings~\cite{peng2018visda,liang2020we,saito2018maximum}. We also use \textbf{COVID-DA}~\cite{zhang2020covid}, adapting the CT images diagnosis from common pneumonia to the novel disease. This is a typical scenario where online domain adaptation is valuable in practice. When a novel disease breaks out, without any prior knowledge, one has to exploit a different but correlated domain to assist the diagnosis of the new pandemic in a time-sensitive manner. We also evaluate on a large-scale medical dataset \textit{Camelyon17} from the \textbf{WILDS}~\cite{koh2021wilds}, a histopathology image datasets with patient population shift from source to the target. Camelyon17 has $455$k samples of breast cancer patients from 5 hospitals. Another practical scenario is the online fashion where the user-generated content (UGC) might be time-sensitive and cannot be saved for training purposes. Due to the lack of cross-domain fashion prediction dataset, we propose to evaluate adapting from \textbf{Fashion-MNIST}~\cite{xiao2017fashion}-to-\textbf{DeepFashion}~\cite{liu2016deepfashion} category prediction branch. We select 6 fashion categories shared between the two datasets, and design the task as adapting from $36,000$ grayscale samples of Fashion-MNIST to $200,486$ real-world commercial samples from DeepFashion.

\begin{table*}[t]
    \footnotesize
    \centering
    \renewcommand{\tabcolsep}{5pt}
    \renewcommand{\arraystretch}{1.2}
     \caption{\small Accuracy on VisDA-C (\%) using ResNet-101. In the online setting, individual class reports accuracy after one-pass, \textit{one-pass} is the class average. Best offline (\textbf{\textit{italic bold}}), best online (\textbf{bold}).
    }\vspace{-0.1in}
    \resizebox{1.0\linewidth}{!}{
    \begin{tabular}{@{}l l c c c c c c c c c c c c c c c@{}}
    \toprule
        \multicolumn{2}{l}{Methods (Syn $\rightarrow$ Real)} & plane & bike & bus & car & horse & knife & motor & person & plant & skate & train & truck & Online & One-pass & Per-Class Acc. \\
        \midrule
        \multirow{5}{*}{Offline} & Source-Only & 67.7 & 27.4 & 50.0 & 61.7 & 69.5 & 13.7 & 85.9 & 11.5 & 64.4 & 34.4 & 84.2 & 19.2 & -  & - & 49.1\\
        {} & DAN~\cite{long2015learning} & 84.4 & 50.9 & 68.4 & 66.8 & 82.0 & 17.0 & 82.3 & 22.0 & 73.3 & 47.4 & 81.2 & 18.3 & - & - & 57.8\\
        {} & CORAL~\cite{sun2016return} & 94.7 & 46.8 & 78.0 & 62.4 & 86.5 & 70.1 & 90.4 & 73.5 & 84.2 & 34.9 & 87.7 & 24.9 & - & - & 69.5\\
        {} & DANN~\cite{ganin2016domain} & 81.9 & 77.7 & \itared{82.8} & 44.3 & 81.2 & 29.5 & 65.1 & 28.6 & 51.9 & 54.6 & 82.8 & 7.8 & -  & - & 57.4\\
        {} & ENT~\cite{saito2019semi} & 88.6 & 29.5 & 82.5 & \itared{75.8} & 88.7 & 16.0 & \itared{93.2} & 63.4 & \itared{94.2} & 40.1 & 87.3 & 12.1& -  & - & 64.3\\
        {} & MDD~\cite{zhang2019bridging} & 89.2 & 58.9 & 70.5 & 54.5 & 71.1 & 42.9 & 78.8 & 22.5 & 68.6 & 54.7 & 88.6 & 15.4& -  & - & 59.6\\
        {} & CDAN~\cite{long2017conditional} & 89.4 & 40.3 & 74.6 & 65.2 & 81.5 & 62.2 & 90.1 & 69.3 & 73.3 & 58.6 & 84.8 & 19.1 & - & - & 67.4\\
        {} & SHOT~\cite{liang2020we} & 94.3 & \itared{88.5} & 80.1 & 57.3 & 93.1 & 94.9 & 80.7 & \itared{80.3} & 91.5 & 89.1 & 86.3 & 58.2 & -  & - & 82.9\\
        {} & ATDOC-NA~\cite{liang2021domain}& \itared{95.3} & 84.7 & 82.4 & 75.6 & \itared{95.8} & \itared{97.7} & 88.7 & 76.6 & 94.0 & \itared{91.7} & \itared{91.5} & \itared{61.9} & -  & - & \itared{86.3}\\
        \midrule
        \multirow{6}{*}{Online} & Source-Only & 73.3 & 6.5 & 44.9 & 67.8 & 58.6 & 5.7 & 67.2 & 18.3 & 47.7 & 19.2 & 84.1 & 9.3 & 46.7 & 41.9 & -\\
        {} & DAN~\cite{long2015learning} & 87.7 & 45.9 & 69.9 & 70.9 & 77.4 & 17.7 & 80.7 & 18.6 & 79.9 & 29.9 & 82.7 & 16.6 & 57.8 & 56.5 & -\\
        {} & CORAL~\cite{sun2016return} & 94.7 & 51.0 & 79.6 & 63.2 & 88.2 & 69.4 & \boblu{91.1} & 73.1 & 87.7 & 41.8 & 88.4 & 24.2 & 66.7 & 71.0 & -\\
        {} & DANN~\cite{ganin2016domain} & 84.5 & 39.2 & 70.2 & 60.4 & 77.1 & 28.6 & 90.9 & 20.5 & 67.7 & 39.9 & \boblu{89.8} & 10.5 & 49.0 & 56.6 & -\\
        {} & ENT~\cite{saito2019semi} & 87.1 & 14.8 & 87.9 & 71.9 & 87.8 & \boblu{98.9} & 90.3 & 0.0 & 5.2 & 15.0 & 80.4 & 0.2 & 55.8 & 53.3 & -\\
        {} & MDD~\cite{zhang2019bridging} & \boblu{95.1} & 52.2 & 87.9 & 57.9 & 90.3 & 94.8 & 88.4 & 45.7 & 76.2 & 50.5 & 77.7 & 25.7 & 60.4 & 70.1 & -\\
        {} & CDAN~\cite{long2017conditional} & 88.5 & 44.3 & 74.3 & 68.4 & 80.3 & 60.2 & 89.9 & 69.9 & 74.3 & 57.1 & 84.8 & 13.9 & 62.3 & 67.1 & -\\
        \midrule
        {} & \textbf{\method{CroDoBo}} & 94.8 & \boblu{87.5} & \boblu{90.5} & \boblu{76.0} & \boblu{94.9} & 93.7 & 88.7 & \boblu{80.1} & \boblu{94.8} & \boblu{89.4} & 84.6 & \boblu{30.7} & \boblu{79.4} & \boblu{84.0} & -\\
         \bottomrule
    \end{tabular}
    }
    \label{tab:visda-c}
    \vspace{-0.1in}
\end{table*}

\noindent \textbf{Implementation details.} We implement using Pytorch~\cite{paszke2017automatic}. We follow~\cite{liang2021domain,liang2020we} to use ResNet-101~\cite{he2016deep} on VisDA-C pretrained on ImageNet~\cite{deng2009imagenet,russakovsky2015imagenet}. We follow~\cite{zhang2020covid} to use pretrained ResNet-18~\cite{he2016deep} on COVID-DA. We follow the leader-board on WILDS challenge~\cite{koh2021wilds}~\footnote{\label{fn:leaderboard}~\url{https://wilds.stanford.edu/leaderboard/}} to use DenseNet-121~\cite{huang2017densely} on Camelyon17 with random initialization, we use the official WILDS codebase~\footnote{~WILDS version: 1.1.0} for data split and evaluation. We use pretrained ResNet-101~\cite{he2016deep} on Fashion-MNIST-to-DeepFashion. The confidence threshold $\tau=0.95$ and diversity weight $\lambda=0.4$ are fixed throughout the experiments.

\noindent\textbf{Baselines.} We compare \method{CroDoBo} with eight state-of-the-art domain adaptation approaches, including \textbf{DAN}~\cite{long2015learning}, \textbf{CORAL}~\cite{sun2016return}, \textbf{DANN}~\cite{ganin2016domain}, \textbf{ENT}~\cite{grandvalet2005semi,saito2019semi}, \textbf{MDD}~\cite{zhang2019bridging}, \textbf{CDAN}~\cite{long2017conditional}, \textbf{SHOT}~\cite{liang2020we} and 
\textbf{ATDOC}~\cite{liang2021domain}. ATDOC has multiple variants of the auxiliary regularizer, we compared with the \textit{Neighborhood Aggregation} (ATDOC-NA) with the best performance in \cite{liang2021domain}. Among the compared approaches, SHOT and ATDOC-NA require a memory module that collects and stores information of all the target samples, thus only apply the offline setting. For the other six approaches, we compare both offline and online results. Each offline model is trained for 10 epochs, and each online model takes the same randomly-perturbed target queries to make a fair comparison.

\begin{figure}[tbh]
    \centering
    \includegraphics[width=0.9\linewidth]{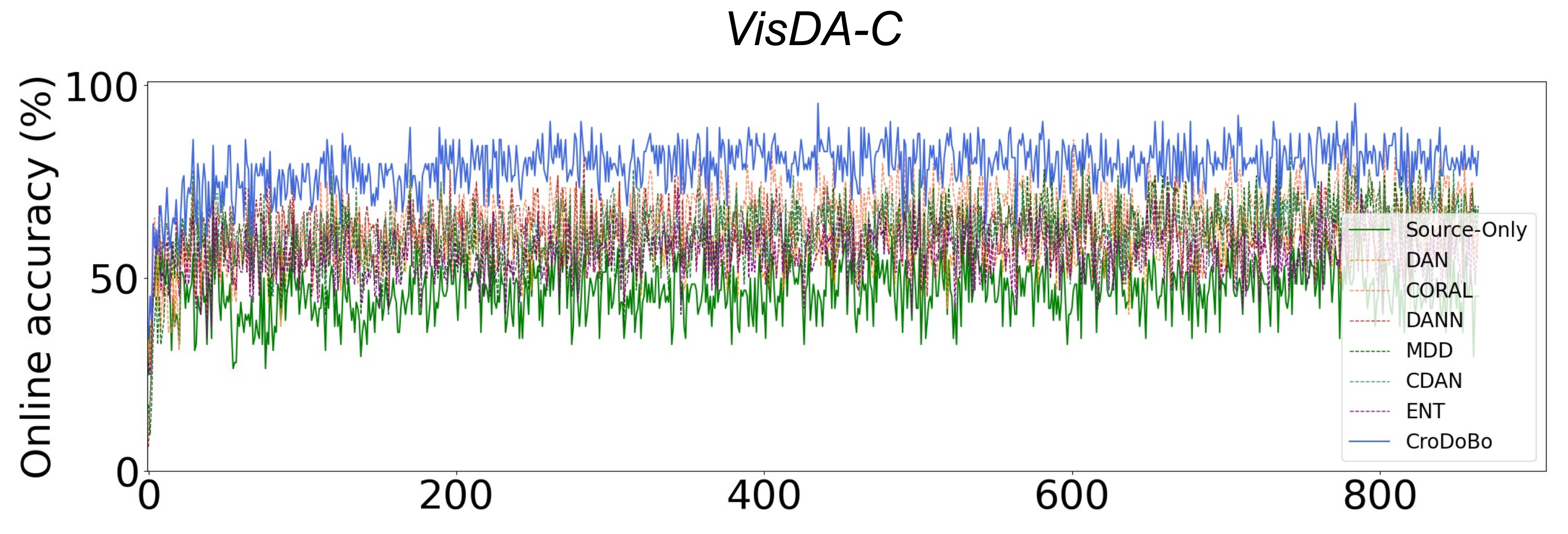}
    \vspace{-0.1in}
    \caption{\small Results of online accuracy on VisDA-C~\cite{peng2018visda}. X-axis is the streaming timestep of the target queries. Each query contains 64 samples. Each approach takes the same randomly perturbed sequence of target queries. Source-Only is the emphasized solid \textcolor{ForestGreen}{green line}, \method{CroDoBo} is the solid \textcolor{Blue}{blue line}.  \textit{Best viewed in color.}}
    \label{fig:visda_acc}
    \vspace{-0.1in}
\end{figure}

\noindent\textbf{Main results.}  We summarize the results on VisDA-C~\cite{peng2018visda} in Table~\ref{tab:visda-c}, and plot the online streaming accuracy in Figure~\ref{fig:visda_acc} To make a fair comparison, we follow~\cite{peng2018visda,kang2019contrastive,liang2020we,liang2021domain} to provide the VisDA-C one-pass accuracy in class average. Among the offline methods, SHOT~\cite{liang2020we} and ATDOC-NA~\cite{liang2020we} largely outperforms other approaches, showing the effectiveness of target domain clustering. Regarding the online setting, the Source-Only baseline loses 2.4\% in the online average and 7.2\% in the one-pass accuracy, which indicates that the data diversity is also important in domain generalization. We observe that ENT~\cite{saito2019semi}, which is an entropy regularizer on the posterior probabilities of the unlabeled target samples, has a noticeable performance drop in the online setting, and illustrates obvious imbalanced results over the categories (superior at class ``knife'' but poor at ``person'' and ``truck''). We consider it a typical example of bad objective choice for the online setting when the dataset is imbalanced. Without sufficient rounds to provide data diversity, entropy minimization might easily overfit the current target query. The 2.5\% drop in one-pass from online further confirmed the model has deviated from the beginning. \method{CroDoBo} outperforms other methods in the online setting by a large margin, and is comparable to the offline result from the state-of-the-art approach ATDOC-NA~\cite{liang2021domain}. For the sake of time efficiency, \method{CroDoBo} is superior than other approaches by achieving high accuracy in only one epoch.

\begin{table}[th]
    \footnotesize
    \centering
    \renewcommand{\tabcolsep}{6pt}
    \renewcommand{\arraystretch}{1.2}
     \caption{\small Accuracy on COVID-DA~\cite{zhang2020covid} (\%) using ResNet-18. COVID-DA* is the method proposed along with dataset in \cite{zhang2020covid}.
    }\vspace{-0.1in}
    \resizebox{0.95\linewidth}{!}{
    \begin{tabular}{@{}l l c c c@{}}
    \toprule
    \multicolumn{2}{l}{Methods (Pneumonia $\rightarrow$ Covid)} &  Online & One-pass & Offline \\
    \midrule
    \multirow{10}{*}{Offline \& Online} & Source-Only & 83.6 & 82.0 & 88.9\\
    {} & DAN~\cite{long2015learning}  & 84.4 & 85.7 & 87.7\\
    {} & CORAL~\cite{sun2016return} & 67.6 & 45.4 & 65.4 \\
    {} & DANN~\cite{ganin2016domain} & 83.0 & 87.1 & 87.7\\
    {} & ENT~\cite{saito2019semi}  & 84.3 & 87.3 & 89.8\\
    {} & MDD~\cite{zhang2019bridging} & 83.2 & 86.2 & 81.0\\
    {} & CDAN~\cite{long2017conditional} & 83.0 & 86.4 & 86.3\\
    {} & SHOT~\cite{liang2020we} & - & - & 93.2\\
    {} & ATDOC-NA~\cite{liang2021domain} & - & - & \itared{98.1}\\
    {} & COVID-DA*~\cite{zhang2020covid}  & - & - &\itared{98.1}\\
    \midrule
    {} & \textbf{\method{CroDoBo}} & \textbf{96.5} & \textbf{97.1} & - \\
    \bottomrule
    \end{tabular}
    }
    \label{tab:covid-da}
    \vspace{-0.2in}
\end{table}

\begin{table}[th]
    \footnotesize
    \centering
    \renewcommand{\tabcolsep}{5pt}
    \renewcommand{\arraystretch}{1.2}
     \caption{\small Accuracy on \textit{WILDS}-Camelyon17~\cite{koh2021wilds} (\%) using DenseNet-121. \textit{Domain Generalization} results are reprinted from \textit{WILDS} leaderboard (see~\cref{fn:leaderboard}).
    }\vspace{-0.1in}
    \resizebox{0.95\linewidth}{!}{
    \begin{tabular}{@{}c l c c c@{}}
    \toprule
    \multicolumn{2}{l}{Methods (Hospital 1,2,3 $\rightarrow$ Hospital 5)} &  Online & One-pass & Offline \\
    \midrule
    {} & ERM~\cite{koh2021wilds} & - & - & 70.3\\
    {} & Group DRO~\cite{koh2021wilds} & - & - & 68.4\\
    {\textit{Domain}} & IRM~\cite{koh2021wilds} & - & - & 64.2\\
    {\textit{Generalization}} & FISH~\cite{shi2021gradient} & - & - & 74.7\\
    \midrule
     \multirow{9}{*}{Offline \& Online} & Source-Only  & 71.7 & 60.1 & 63.6\\
    {} & DAN~\cite{long2015learning}  & 76.3 & 78.0 & 69.0\\
    {} & CORAL~\cite{sun2016return} & 66.0 & 87.1 & 85.0 \\
    {} & DANN~\cite{ganin2016domain} & 76.4 & 81.4 & 86.7\\
    {} & ENT~\cite{saito2019semi}  & 83.1 & 82.3 & \itared{87.5}\\
    {} & MDD~\cite{zhang2019bridging} & 77.8 & 52.5 & 63.7\\
    {} & CDAN~\cite{long2017conditional} & 62.7 & 60.1 & 58.5\\
    {} & SHOT~\cite{liang2020we} & - & - & 73.8\\
    {} & ATDOC-NA~\cite{liang2021domain} & - & - & 86.3\\
    \midrule
    {} & \textbf{\method{CroDoBo}} & \textbf{89.2} & \textbf{91.9} & -\\
    \bottomrule
    \end{tabular}
    }
    \label{tab:camelyon17}
    \vspace{-0.1in}
\end{table}

\begin{table}[th]
    \footnotesize
    \centering
    \renewcommand{\tabcolsep}{5pt}
    \renewcommand{\arraystretch}{1.2}
     \caption{\small Accuracy on Fashion-MNIST~\cite{xiao2017fashion} to DeepFashion~\cite{liu2016deepfashion} (\%) using ResNet-101.
    }\vspace{-0.1in}
    \resizebox{0.95\linewidth}{!}{
    \begin{tabular}{@{}l l c c c@{}}
    \toprule
    \multicolumn{2}{l}{Methods (F-MNIST $\rightarrow$ DeepFashion)} &  Online & One-pass & Offline \\
    \midrule
    \multirow{8}{*}{Offline \& Online} & Source-Only  & 22.7 & 15.8 & 23.1\\
    {} & DAN~\cite{long2015learning}  & 40.7 & 42.0 & 32.7\\
    {} & CORAL~\cite{sun2016return} & 40.4 & 40.7 & 39.6\\
    {} & DANN~\cite{ganin2016domain} & 35.6 & 26.5 & 40.5\\
    {} & ENT~\cite{saito2019semi}& 31.9 & 31.2 & 31.1\\
    {} & MDD~\cite{zhang2019bridging} & 36.5 & 38.0 & 39.0\\
    {} & CDAN~\cite{long2017conditional} & 45.4 & \textbf{47.6} & 47.2\\
    {} & SHOT~\cite{liang2020we} & - & - & 42.3\\
    {} & ATDOC-NA~\cite{liang2021domain} & - & - & \itared{47.4}\\
    \midrule
    {} & \textbf{\method{CroDoBo}} & \textbf{49.1} & 46.3 & -\\
    \bottomrule
    \end{tabular}
    }
    \label{tab:fashion}
    \vspace{-0.1in}
\end{table}

\begin{figure*}[t]
    \centering
    \includegraphics[width=0.95\linewidth]{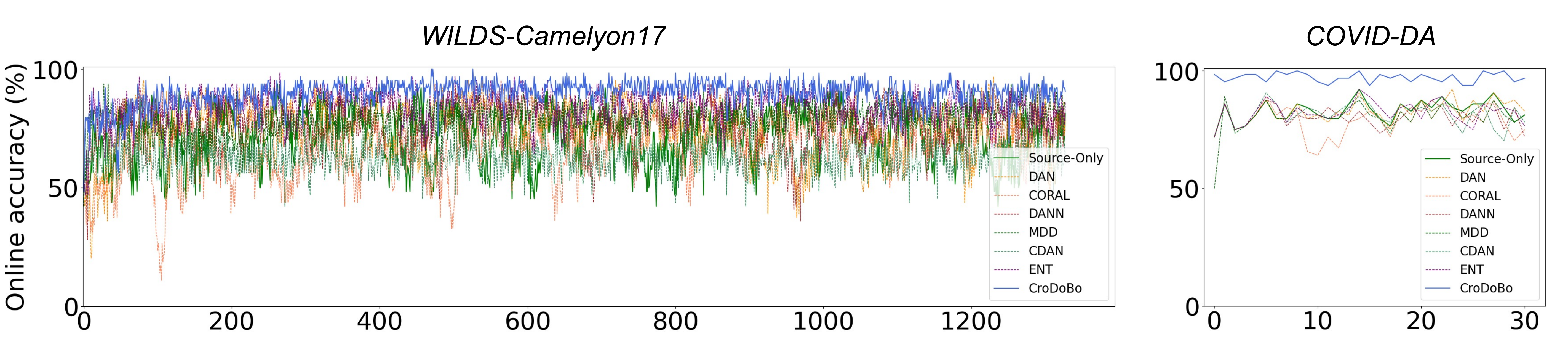}
    \vspace{-0.1in}
    \caption{\small Results of online accuracy on \textit{WILDS}-Camelyon17~\cite{koh2021wilds}, COVID-DA~\cite{zhang2020covid}. Each query contains 64 samples. Source-Only is the emphasized solid \textcolor{ForestGreen}{green line}, \method{CroDoBo} is the solid \textcolor{Blue}{blue line}. \textit{Best viewed in color.}}
    \label{fig:medical_acc}
    \vspace{-0.1in}
\end{figure*}

\begin{figure*}[t]
    \centering
    \includegraphics[width=1.0\linewidth]{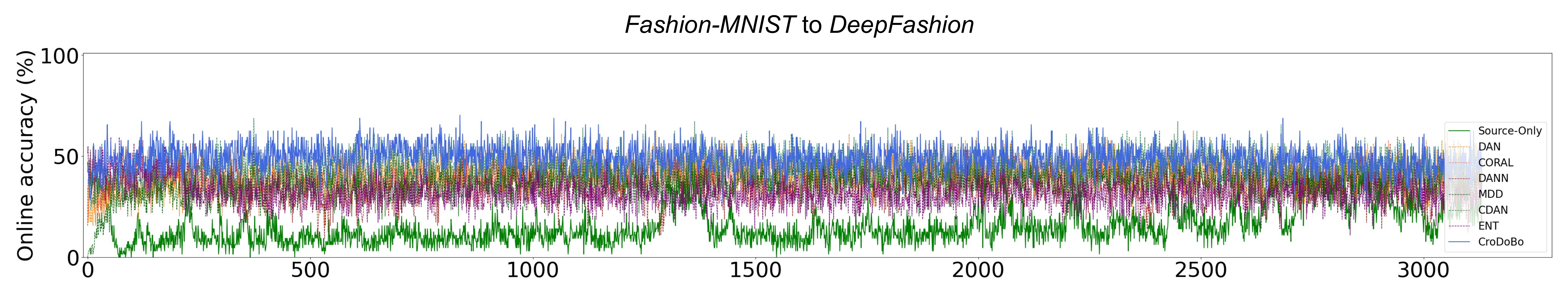}
    \vspace{-0.2in}
    \caption{\small Results of online accuracy on Fashion-MNIST-to-DeepFashion~\cite{xiao2017fashion,liu2016deepfashion}. Each query contains 64 samples. \textit{Best viewed in color.}}
    \label{fig:fashion_acc}
    \vspace{-0.2in}
\end{figure*}

\begin{figure*}[t]
    \centering
    \includegraphics[width=1.0\linewidth]{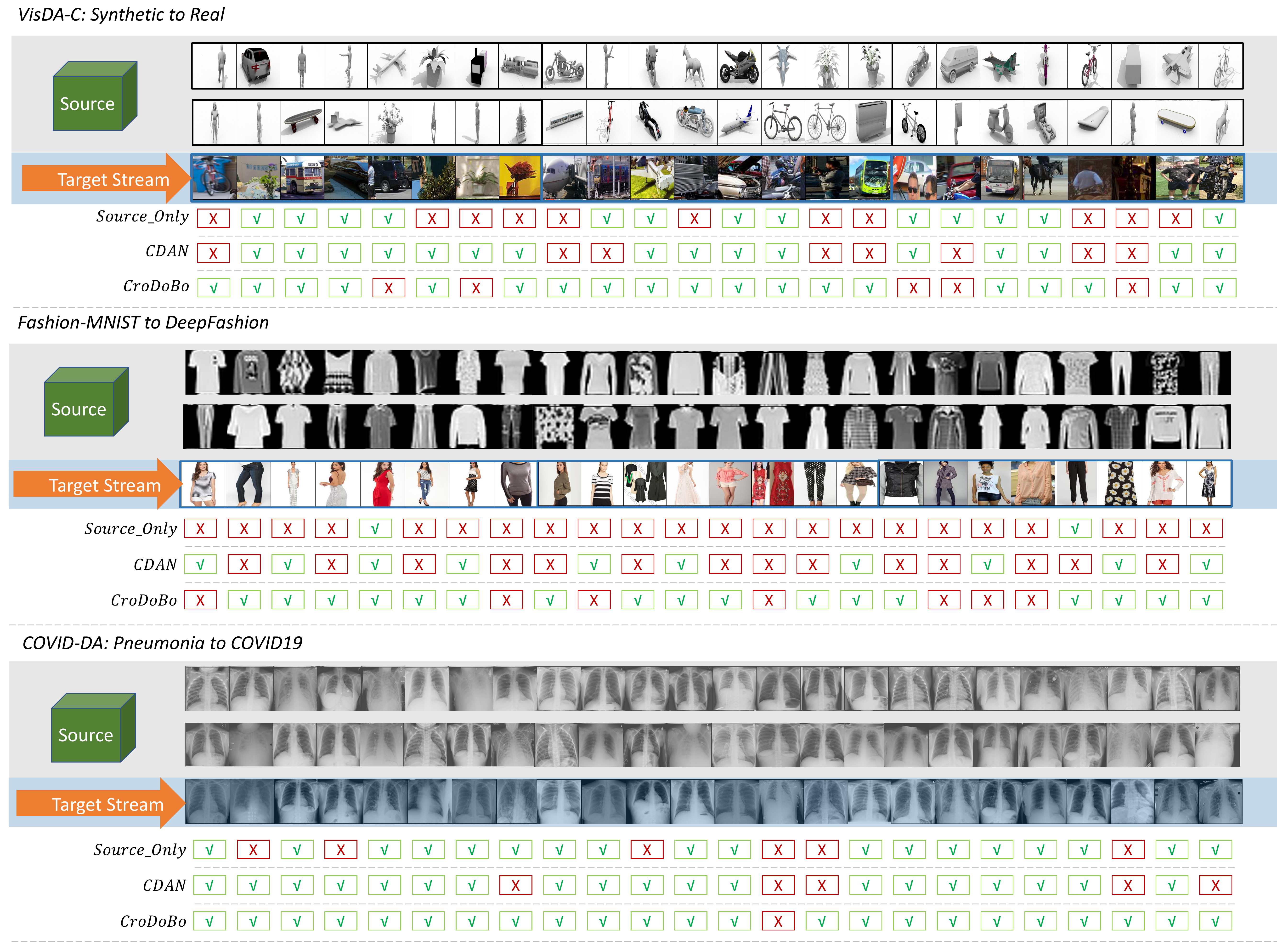}
    \vspace{-0.2in}
    \caption{\small Qualitative results of a randomly selected target query (size 24). We compare \method{CroDoBo} with two essential baselines Source-Only and CDAN~\cite{long2017conditional}. We represent the bootstrapped source samples (top two rows under each benchmark), target samples (third row under each benchmark), and the prediction result of each target sample. \textit{Best viewed in color.}}
    \label{fig:qualitative}
    \vspace{-0.2in}
\end{figure*}

The results on two medical imaging datasets COVID-DA~\cite{zhang2020covid} and \textit{WILDS}-Camelyon17~\cite{koh2021wilds} are respectively summarized in Table~\ref{tab:covid-da} and Table~\ref{tab:camelyon17}.The online streaming accuracy is presented in Figure~\ref{fig:medical_acc}. COVID-DA* is the method proposed along with the dataset in~\cite{zhang2020covid}, which is a domain adversarial-based multi-classifier approach with focal loss regularization. \method{CroDoBo} outperforms the other approaches on COVID-DA regarding the online and one-pass metric, and achieves competitive performance against the best offline accuracy. On the large-scale benchmark \textit{WILDS}-Camelyon17, \method{CroDoBo} outperforms the offline results by 1.7\%, which validates the effectiveness of the approach. We notice that \method{CroDoBo} obtains more competitive results on large-scale datasets than small ones. Compared to the COVID-DA dataset (2,029 target samples), \textit{WILDS}-Camelyon17 (85,054 target samples) is significantly more extensive. The good performance on larger number of target queries indicates that \method{CroDoBo} can well exploit the underlying information from the target domain. Similar observations are made on the large-scale Fashion benchmark~\cite{xiao2017fashion,liu2016deepfashion}. Meanwhile, we reprint \textit{Domain Generalization} results from the \textit{WILDS} leaderboard just for reference. Despite bearing a different setting, the online adaptation does not overwhelm the system with much more computational cost, but can improve the precision of the model by a large margin.

The results on the newly proposed large-scale Fashion benchmark, from Fashion-MNIST~\cite{xiao2017fashion} to DeepFashion~\cite{liu2016deepfashion} category prediction branch, is summarized in Table~\ref{tab:fashion}. We also plot the online streaming accuracy in Figure~\ref{fig:fashion_acc}. To the best of our knowledge, we are the first to report results on this adaptation scenario. The offline Source-Only merely achieves 23.1\% accuracy, only 6.5\% gain on the basis of the probability of guessing, which indicates the benchmark is challenging. The sharp drop of performance from Source-Only online accuracy to one-pass accuracy (-6.8\%) indicates the large domain gap, and how easy the model is dominated by the source domain supervision. Similar observation is made on \textit{WILDS}-Camelyon17 Source-Only results(-11.6\% from online to one-pass), this usually happens when the source domain is less challenging than the target domain, and the distribution of the two domains are far from each other. Faced with this challenging benchmark, \method{CroDoBo} improves the online performance to a remarkable 49.1\%, outperforming the best result in the offline setting. Our one-pass accuracy is slightly shy compared to CDAN~\cite{long2017conditional}, but is better in online metric.

\noindent\textbf{Ablation study.}
We conduct ablation study on the impact of cross-domain bootstrapping in Table~\ref{tab:ablation-crodobo}. Following Table~\ref{tab:visda-c}, we provide the VisDA-C one-pass accuracy in class average. This study is to evaluate whether the improvement is introduced by cross-domain bootstrapping or simply the strong baseline with the objectives on the target domain (see ~\cref{sec:exchange}). Thus, we devise a baseline by removing only the cross-domain bootstrapping, called w/o \method{CroDoBo}. The baseline model has one learner that is optimized by minimizing the objective
$\ell_s + \ell_t + \ell_\text{ent} + \lambda\ell_\text{div}$, where $\ell_t=B^{-1}\sum_{b=1}^{B} \mathbb{1} \left(p_b \geq \tau\right) \mathcal{H}\left(\hat{y}_b;p(c|\Tilde{t}_b;\vw)\right)$, which is ~\cref{eq:fixmatch} without exchanging the pseudo-labels.
In Table~\ref{tab:ablation-crodobo}, we observe that w/ \method{CroDoBo} is consistently better than w/o in the online average accuracy on all the datasets. Regarding one-pass accuracy, the effectiveness of cross-domain bootstrapping is unapparent on smaller datasets VisDA-C and COVID-DA, yet clearly outperforms w/o on large-scale \textit{WILDS}-Camelyon17 and Fashion-MNIST-to-DeepFashion.

\begin{table}[t]
    \footnotesize
    \centering
    \renewcommand{\tabcolsep}{4pt}
    \renewcommand{\arraystretch}{1.2}
     \caption{\small Ablation study of cross-domain bootstrapping on four datasets (\%). VisDA-C one-pass accuracy is in per-class.
    }\vspace{-0.1in}
    \resizebox{1.0\linewidth}{!}{
    \begin{tabular}{@{}l l c c c c@{}}
    \toprule
    \multicolumn{2}{l}{Method/Dataset} & VisDA-C & COVID-DA & Camelyon17 & Fashion \\
    \midrule
    \multirow{2}{*}{Online} & w/o CroDoBo & 78.5 & 94.4 & 86.2 & 42.3\\
    {} & w/ CroDoBo & 79.4 &  96.5 & 89.2 & 49.1\\
    \midrule
    \multirow{2}{*}{One-pass} & w/o CroDoBo & 84.0 & 97.1 & 89.4 & 39.9\\
    {} & w/ CroDoBo & 84.0 & 97.1 & 91.9 & 46.3\\
    \bottomrule
    \end{tabular}
    }
    \label{tab:ablation-crodobo}
    \vspace{-0.2in}
\end{table}

\begin{table}[th]
    \footnotesize
    \centering
    \renewcommand{\tabcolsep}{6pt}
    \renewcommand{\arraystretch}{1.2}
     \caption{\small Ablation study on the objectives on target domain on VisDA-C (\%). $T$ is the sharpening temperature in the MixMatch~\cite{berthelot2019mixmatch} algorithm. VisDA-C one-pass accuracy is in per-class.}\vspace{-0.1in}
    \resizebox{1.0\linewidth}{!}{
    \begin{tabular}{@{}l c c@{}}
    \toprule
    Method & Online & One-pass\\
    \midrule
    \textbf{default} (w/o \method{CroDoBo}, $\tau$=0.95, $\lambda$=0.4) & 78.5 (-) & 84.0 (-)\\
    w/o~~$\ell_\text{ent}$ & 63.7(\down) & 53.1(\down)\\ 
    w/o~~$\ell_\text{div}$ & 72.6(\down) & 73.0(\down)\\
    replace $\ell_\text{ent} + \ell_\text{div}$ w/ Pseudo-labeling~\cite{lee2013pseudo} ($\tau$=0.95) & 70.2(\down) & 70.0(\down)\\
    replace $\ell_\text{ent} + \ell_\text{div}$ w/ MixMatch~\cite{berthelot2019mixmatch} ($T$=0.5) & 73.0(\down) & 75.3(\down)\\
    replace $\ell_t$ w/ MixMatch~\cite{berthelot2019mixmatch} ($T$=0.5) & 76.3(\down) & 81.5(\down)\\
    use \textit{Randaug}~\cite{cubuk2020randaugment} on $\ell_\text{ent}, \ell_\text{div}$ & 77.6(\down) & 83.7(\down)\\
    \bottomrule
    \end{tabular}
    }
    \label{tab:ablation_target}
    \vspace{-0.3in}
\end{table}

We further conduct ablation study on the objective terms (see~\cref{sec:exchange}) and report the results in Table~\ref{tab:ablation_target}. To eliminate the benefit of cross-domain boosting, our default setting is the model w/o \method{CRODOBO}. We leave out $\ell_\text{ent}$ and observe significant performance drop. Without $\ell_\text{div}$, the performance decrease slight in the online metric, but far more sharply on the one-pass metric (which is calculated per-class). We analyze that the diversity term is important for imbalanced dataset like VisDA-C to achieve high class-average accuracy. We also report the results by replacing $\ell_\text{ent}$ and $\ell_\text{div}$ with Pseudo-labeling~\cite{lee2013pseudo}.
We replace either $\{\ell_\text{ent},\ell_\text{div}\}$ or $\ell_t$ with MixMatch, and observe decent performance when employed together with $\{\ell_\text{ent},\ell_\text{div}\}$ (see Table~\ref{tab:ablation_target} row6). The RandAugment~\cite{cubuk2020randaugment} on the entropy and diversity terms does not enhance the performance.
\vspace{-0.1in}
\section{Conclusion}
In the context of the \emph{the right to be forgotten}, we propose an online domain adaptation framework in which the target data is erased immediately after prediction. A novel online UDA algorithm is proposed to tackle the lack of data diversity, which is a fundamental drawback of the online setting. The proposed method achieves state-of-the-art online results and comparable results to the offline domain adaptation approaches. We would like to extend \method{CroDoBo} to more tasks like semantic segmentation~\cite{li2019bidirectional,zhao2019multi}.

\noindent\textbf{Potential negative impact.} The proposed method is at the risk of privacy leakage if one exploits purposefully the memorization effect of the deep neural network weights and restore the private information, which is the common vulnerability of all neural networks.

\section*{Acknowledgement}
We thank Caiming Xiong for his valuable insights into the project, Junnan Li and Shu Zhang for the help to improve the experiments.

{\small
\bibliographystyle{ieee_fullname}
\bibliography{egbib}
}

\appendix
\section*{Supplementary Material}
\section{Prior Online UDA approaches}
\label{prior_online}
In the main paper, we propose a novel cross-domain framework to implement \emph{the right to be forgotten}. However, we \emph{do not} claim to have proposed the task of online unsupervised domain adaptation, which has existed before the emergence of deep learning~\cite{dredze2008online,jain2011online,moon2020multi}. The more recent works are either engineered for a specific task that lacks generality~\cite{delussu2021online,gaidon2015online,wu2016online,mancini2018kitting,hajifar2021online} or more general to compare to but not published~\cite{taufique2021conda}. Yet, we try to compare to the unpublished approach~\cite{taufique2021conda} despite its limited availability. The setting of~\cite{taufique2021conda} is different from our approach, with a memory module that buffers some previous target queries that the model can re-access, ~\cite{taufique2021conda} is certainly less challenging comparing to ``\textit{burn after reading}''. Meanwhile, ~\cite{taufique2021conda} bears a continual setting, in which the model is pretrained on the source domain and then adapted to the target domain. Without source code, we compare to the results they present in the paper with the same employed backbone \textit{HR-Net}~\cite{wang2020deep}. We devise \method{CroDoBo} to a continual setting to make it comparable. Without simultaneous access to the source domain, cross-domain bootstrapping is not an option. So we employ the objectives on the target domain (see main paper Sec.\textcolor{red}{3.2.2}), we call it \emph{Continual \method{CroDoBo}}.
The comparison results are in Table~\ref{tab:compare_ConDA}.

We observe that, without any buffer mechanism or re-access to the previous queries, the continual \method{CroDoBo} still outperforms ConDA~\cite{taufique2021conda}.

\begin{table*}[t]
    \footnotesize
    \centering
    \renewcommand{\tabcolsep}{5pt}
    \renewcommand{\arraystretch}{1.2}
     \caption{\small Accuracy on VisDA-C (\%) using HR-Net.
    }\vspace{-0.1in}
    \resizebox{1.0\linewidth}{!}{
    \begin{tabular}{@{}l c c c c c c c c c c c c c c c@{}}
    \toprule
        Methods (Syn $\rightarrow$ Real) & plane & bike & bus & car & horse & knife & motor & person & plant & skate & train & truck & Online & One-pass \\
        \midrule
        ConDA~\cite{taufique2021conda} & 97.0 & 90.4 & 80.9 & 50.0 & 95.2 & 95.7 & 80.3 & 81.9 & 94.9 & 94.2 & 91.1 & 63.9 & N/P & 84.6 \\
        Continual \method{CroDoBo} (Ours) &  96.5 & 85.2 & 82.3 & 47.3 & 98.0 & 96.1 & 89.6 & 79.2 & 94.9 & 95.7 & 90.4 & 66.5 & 80.0 & 85.1\\
        \textbf{\method{CroDoBo}} & 94.8 & 86.0 & 90.7 & 80.3 & 97.1 & 99.1 & 93.1 & 85.0 & 88.2 & 89.6 & 90.9 & 47.1 & 82.9 & 86.8\\
         \bottomrule
    \end{tabular}
    }
    \label{tab:compare_ConDA}
    \vspace{-0.1in}
\end{table*}

\section{Streaming Randomness}
\label{randomness}
As mentioned in the main paper (Sec.\textcolor{red}{4}), each online model takes the same randomly-perturbed target queries. Here, we discuss whether the performance of the model can be influenced by different random sequential orders. We perturb the original target sequence (arrange in the order of category) using 5 different random seeds, and report the results of each seed on VisDA-C~\cite{peng2018visda} and the large-scale Fashion-MNIST-to-DeepFashion~\cite{liu2016deepfashion} benchmark. We compare the randomness using CDAN~\cite{long2017conditional} and \method{CroDoBo}. We choose CDAN~\cite{long2017conditional} since it is the state-of-the-art adversarial approach and is essentially different from our proposed \method{CroDoBo}. The results are in Table~\ref{tab:randomness}. We observe that on VisDA-C the variance among different sequential orders is rather small ($<$ 0.25). On the more challenging Fashion benchmark, the variance of \method{CroDoBo} is larger but manageable ($<$ 2.0). We analyze that \method{CroDoBo} relies more on the target-oriented supervision (see main paper Sec.\textcolor{red}{3.2.2}) than CDAN~\cite{long2017conditional}, which makes it more sensitive towards the changes of the target samples. This is a drawback of \method{CroDoBo} that we will try to address in the future work.

To conclude, the randomness in forming the order of target queries will not influence the evaluation on the method's effectiveness.
\begin{table*}[h]
    \footnotesize
    \centering
    \renewcommand{\tabcolsep}{5pt}
    \renewcommand{\arraystretch}{1.2}
     \caption{\small Replacing main paper Eq.\textcolor{red}{3} with other pseudo-labeling methods(\%) on ViDA-C.
    }\vspace{-0.1in}
    \resizebox{1.0\linewidth}{!}{
    \begin{tabular}{@{}l c c c c c c c c c c c c c c@{}}
    \toprule
        Methods (Syn $\rightarrow$ Real) & plane & bike & bus & car & horse & knife & motor & person & plant & skate & train & truck & Online \\
        \midrule
        $w^u$: MixMatch~\cite{berthelot2019mixmatch} $w^v$: FixMatch~\cite{sohn2020fixmatch} & 93.2 & 80.9 & 85.6 & 67.1 & 94.1 & 10.3 & 88.4 & 77.9 & 92.3 & 91.9 & 85.7 & 35.9 & 74.3\\
        $w^u$, $w^v$: MixMatch~\cite{berthelot2019mixmatch} & 94.7 & 83.3 & 81.0 & 62.4 & 90.7 & 13.8 & 84.8 & 78.7 & 95.6 & 94.6 & 82.9 & 45.4 & 71.6\\
        \textbf{\method{CroDoBo}} & 94.8 & 87.5 & 90.5 & 76.0 & 94.9 & 93.7 & 88.7 & 80.1 & 94.8 & 89.4 & 84.6 & 30.7 & 79.4\\
         \bottomrule
    \end{tabular}
    }
    \label{tab:other_pl}
    \vspace{-0.1in}
\end{table*}

\begin{table}[thb]
    \footnotesize
    \centering
    \renewcommand{\tabcolsep}{6pt}
    \renewcommand{\arraystretch}{1.2}
     \caption{\small Online accuracy (\%) on five different perturbations of target sequence on VisDA-C~\cite{peng2018visda} and Fashion-MNIST~\cite{xiao2017fashion}-to-DeepFashion~\cite{liu2016deepfashion}.
    }\vspace{-0.1in}
    \resizebox{1.0\linewidth}{!}{
    \begin{tabular}{@{}l c c c c c c c@{}}
    \toprule
    \multicolumn{7}{c}{VisDA-C}\\
    \midrule
    Methods & rand 0 & rand 1 & rand 2 & rand 3 & rand 4 & mean & var\\
    \midrule
    CDAN~\cite{long2017conditional} & 62.3 & 61.0 & 61.9 & 61.6 & 61.9 & 61.7 & 0.21\\
    \method{CroDoBo} & 79.4 & 78.6 & 79.6 & 79.2 & 79.4 & 79.2 & 0.15\\
    \midrule
    \multicolumn{7}{c}{Fashion-MNIST-to-DeepFashion}\\
    \midrule
    Methods & rand 0 & rand 1 & rand 2 & rand 3 & rand 4 & mean & var\\
    CDAN~\cite{long2017conditional} & 45.4 & 47.4 & 46.7 & 46.3 & 46.2 & 46.4 & 0.54\\
    \method{CroDoBo} & 49.1 & 48.9 & 46.3 & 46.5 & 48.9 & 47.9 & 1.99\\
    \bottomrule
    \end{tabular}
    }
    \label{tab:randomness}
\end{table}

\section{Other Pseudo-labeling Approaches as Co-supervision}
\label{other_pl}
The co-supervision in the proposed method (cf. main paper Eq.\textcolor{red}{(3)}) can be replaced with any other pseudo-labeling approaches. With the constant proposal of new pseudo-labeling techniques, one can simply replace the term on either/both \{$w^u$, $w^v$\} to achieve better performance. We replace on either/both learners with another popular semi-supervised approach MixMatch~\cite{berthelot2019mixmatch} and report the results in Table~\ref{tab:other_pl}. We observe that FixMatch~\cite{sohn2020fixmatch} provides better co-supervision and the online performance drops $\sim$8\% when replaced with MixMatch.

\begin{table}[h]
    \footnotesize
    \centering
    \renewcommand{\tabcolsep}{6pt}
    \renewcommand{\arraystretch}{1.2}
     \caption{\small Performance sensitivity (\%) to hyperparameter $\lambda$ (weight for diversity loss in main paper Eq.\textcolor{red}{(4)}) on VisDA-C~\cite{peng2018visda}, $\tau$=0.95.}\vspace{-0.1in}
    \resizebox{1.0\linewidth}{!}{
    \begin{tabular}{@{}l c c c c c c c@{}}
    \toprule
    Metric/$\lambda$ & 0.1 & 0.4 & 0.5 & 0.8 & 1.0 & mean & var\\
    \midrule
    Online & 74.9 & 79.4 & 78.7 & 78.5 & 78.4 & 78.0 & 3.1\\
    One-pass & 80.2 & 84.0 & 83.4 & 83.6 & 83.5 & 82.9 & 2.4\\
    \bottomrule
    \end{tabular}
    }
    \label{tab:hyper_lambda}
\end{table}

\begin{table}[h]
    \footnotesize
    \centering
    \renewcommand{\tabcolsep}{6pt}
    \renewcommand{\arraystretch}{1.2}
     \caption{\small Performance sensitivity (\%) to hyperparameter $\tau$ (confidence threshold for pseudo-label selection in main paper Eq.\textcolor{red}{(3)}) on VisDA-C~\cite{peng2018visda}, $\lambda$=0.4.}\vspace{-0.1in}
    \resizebox{1.0\linewidth}{!}{
    \begin{tabular}{@{}l c c c c c c c c@{}}
    \toprule
    Metric/$\tau$ & 0.5 & 0.6 & 0.7 & 0.8 & 0.9 & 0.95 & mean & var\\
    \midrule
    Online & 75.0 & 76.7 & 77.3 & 77.9 & 78.4 & 79.4 & 77.5 & 2.3\\
    One-pass & 80.9 & 81.7 & 82.6 & 82.8 & 83.4 & 84.0 & 82.7 & 2.0\\
    \bottomrule
    \end{tabular}
    }
    \label{tab:hyper_tau}
\end{table}

\begin{figure}[h]
    \centering
    \includegraphics[width=1.0\linewidth]{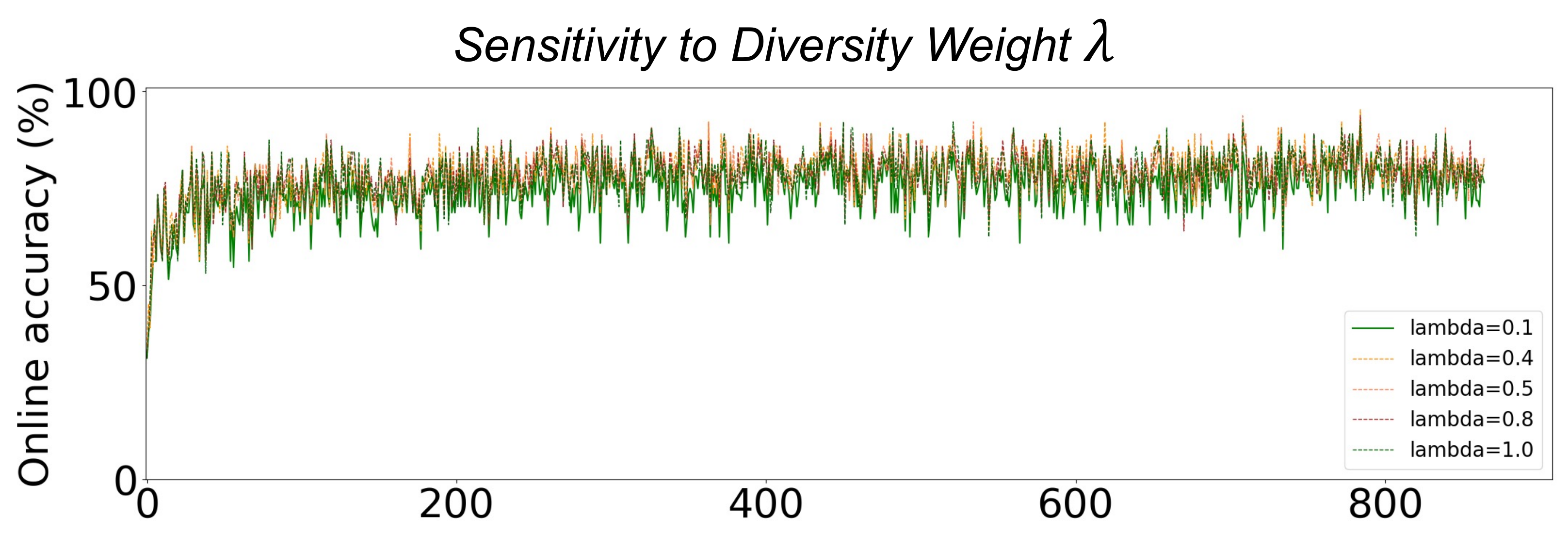}
    \caption{Results of online accuracy \textit{w.r.t.} sensitivity of hyperparameter $\lambda$ on VisDA-C~\cite{peng2018visda}.}
    \label{fig:sensitivity_lambda}
\end{figure}

\begin{figure}[h]
    \centering
    \includegraphics[width=1.0\linewidth]{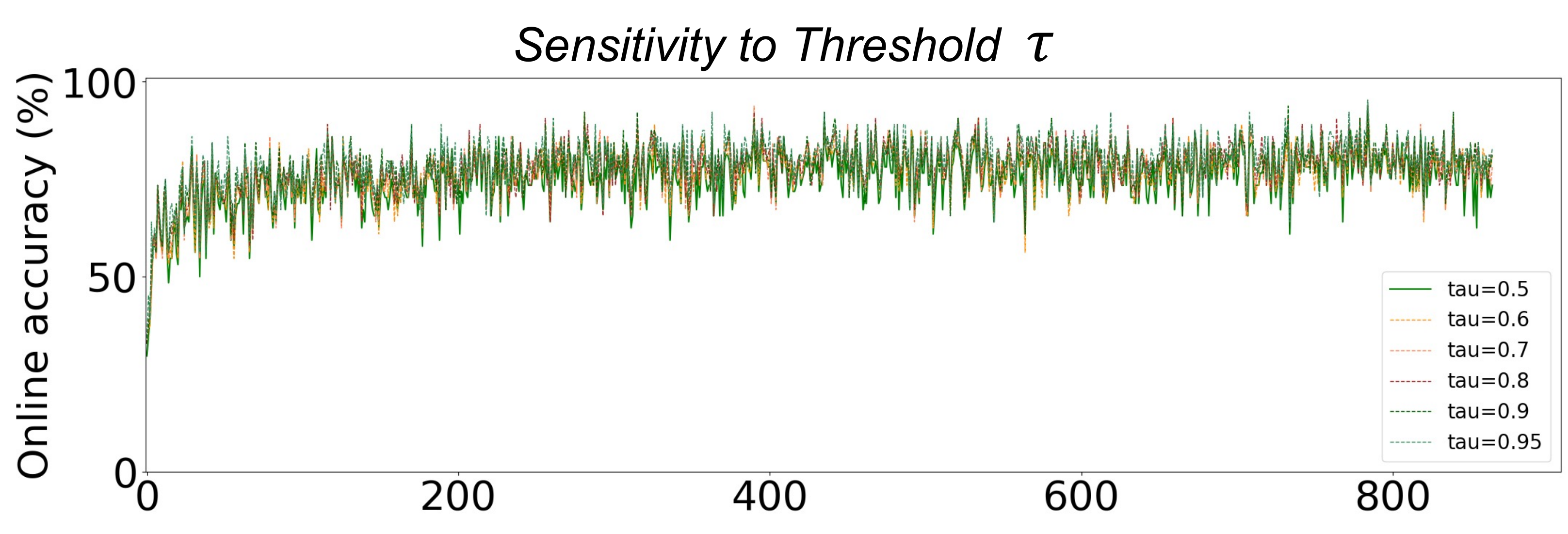}
    \caption{Results of online accuracy \textit{w.r.t.} sensitivity of hyperparameter $\tau$ on VisDA-C~\cite{peng2018visda}.}
    \label{fig:sensitivity_tau}
\end{figure}

\section{Hyperparameters}
\label{hyper}
We have two hyperparameters in the proposed approach: $\lambda$ for weighing the term $\ell_\text{div}$ and $\tau$ for the pseudo-label selection (cf. main paper Eq.\textcolor{red}{(3)}). We used $\lambda$=0.4 and $\tau$=0.95 in all our experiments, here we report results on more settings of these hyperparameters. The results of $\lambda$=\{0.1, 0.4, 0.5, 0.8, 1.0\} are shown in Table~\ref{tab:hyper_lambda}. As the results suggest, \method{CroDoBo} is not sensitive to hyperparameter $\lambda$. We observe similar performance of the model when $\lambda$ is larger than 0.4.

The sensitivity to $\tau$ is shown in Table~\ref{tab:hyper_tau}. When $\tau$ is smaller, more samples in each target query are selected as pseudo-labels to co-supervise the peer learner. However, the quality of these pseudo-labels is compromised since the model is less confident about the prediction. Thus, the co-supervision is less accurate to depend on. We observe the performance drop when the threshold $\tau$ is smaller than 0.6. Therefore, we suggest a larger threhold $\tau$ to achieve a more effective model. The online accuracy of the above settings are shown in Figure~\ref{fig:sensitivity_lambda} and Figure~\ref{fig:sensitivity_tau}.

\section{Network Architecture}
\label{network_detail}
We follow the network architecture in~\cite{liang2020we,liang2021domain}, a feature backbone followed by a bottleneck layer with dimension=256, and a Linear layer as the output layer. For the experiments on VisDA-C~\cite{peng2018visda}, COVID-DA~\cite{zhang2020covid} and Fashion-MNIST-to-DeepFashion~\cite{xiao2017fashion,liu2016deepfashion}, the feature backbone is pretrained on ImageNet~\cite{deng2009imagenet}. For the \textit{WILDS}-Camelyon17 benchmark, we followed the leaderboard to use a randomly initialized DenseNet-121~\cite{huang2017densely}. We use Adam~\cite{kingma2014adam} with with an initial learning rate of 8e-4. The query size in our experiments is set as 64. We have not observed any major performance change using different batch-size.

\end{document}